# SAT2LOD2: A SOFTWARE FOR AUTOMATED LOD-2 BUILDING RECONSTRUCTION FROM SATELLITE-DERIVED ORTHOPHOTO AND DIGITAL SURFACE MODEL

Shengxi Gui [1,2], Rongjun Qin [1,2,3,4] *, Yang Tang[1,2]

[1] Geospatial Data Analytics Laboratory, The Ohio State University, Columbus, USA
[2] Department of Civil, Environmental and Geodetic Engineering, The Ohio State University, Columbus, USA
[3] Department of Electrical and Computer Engineering, The Ohio State University, Columbus, USA
[4] Translational Data Analytics Institute, The Ohio State University, Columbus, USA
Email: <gui.55, qin.324, tang.1693> @osu.edu

**Commission II, WG II/4**

**ABSTRACT:**

Deriving LoD2 models from orthophoto and digital surface models (DSM) reconstructed from satellite images is a challenging task. Existing solutions are mostly system approaches that require complicated step-wise processes, including not only heuristic geometric operations, but also high-level steps such as machine learning-based semantic segmentation and building detection. Here in this paper, we describe an open-source tool, called SAT2LOD2, built based on a minorly modified version of our recently published work. SAT2LoD2 is a fully open-source and GUI (Graphics User Interface) based software, coded in Python, which takes an orthophoto and DSM as inputs, and outputs individual building models, and it can additionally take road network shapefiles, and customized classification maps to further improve the reconstruction results. We further improve the robustness of the method by 1) intergrading building segmentation based on HRNetV2 into our software; and 2) having implemented a decision strategy to identify complex buildings and directly generate mesh to avoid erroneous LoD2 reconstruction from a system point of view. The software can process a moderate level of data (around 5000*5000 size of orthophoto and DSM) using a PC with a graphics card supporting CUDA. Furthermore, the GUI is self-contained and stores the intermediate processing results facilitating researchers to learn the process easily and reuse intermediate files as needed. The updated codes and software are available under this GitHub page: https://github.com/GDAOSU/LOD2BuildingModel.

## 1. INTRODUCTION

### 1.1 Background and related work

Level-of-Detail (LoD) building model is a typical definition for building geometric 3D models from city geography markup language (CityGML) ranging from 0 to 4 (Gröger et al., 2008) and sub-level for detailed models (Gröger et al., 2012; Biljecki et al., 2016). 3D landscape models derived from satellite imagery have the unique advantage to cover wide-area and being low-cost, hence satellite data can be a significant source for building modeling. Therefore, with fine DSM derived from satellite imagery, Level-of-Detail 1 (LoD-1) or higher-level 3D building model can be generated by building modeling approaches. Nevertheless, most current methods remain the challenges to handle 3D building reconstruction at the LoD-2 level.

Typical building model reconstruction methods from remotely sensed data are divided into bottom-up strategies and top-down strategies. The bottom-up strategy (data-driven strategy) assumes building to be individual parts of roof planes, and then reconstruct the structure by the geometrical relationship of point, lines, and surface from DSMs and point cloud. The structure of the roof plane and roof boundary can be achieved by using feature filling (Zhou et al., 2016), or region growing (Sun & Salvaggio, 2013). Top-down strategy (model-driven strategy), predefine a 3D building model library and then fulfill the model with the best parameters based on the DSM or point cloud (Lafarge et al., 2008, Huang et al., 2013). Building model reconstruction using image products often comes with complicated steps following heuristic rules to approximate the topology of man-made architectures. Therefore, existing approaches are mostly system approaches requiring complex implementation steps or numerous tunable parameters to adapt to the variable building landscapes (Yu et al., 2021; Brédif et al., 2013; Bonczak & Kontokosta, 2019). Other approaches based on planar arrangements (Bauchet & Lafarge, 2020) and data optimization (Nan & Wonka, 2017; Shah et al., 2017) have less heuristic processing, while these approaches are designed for very dense point clouds, and satellite-based data do not cover sufficient 3D points for each building to reconstruct accurate 3D building model. A typical building model generation pipeline based on nDSM and the panchromatic image starts from building mask detection, and then building 2D parameters extraction, and 3D parameters are computed by specific model fitting algorithms (Alidoost et al., 2019; Partovi et al., 2019). Some of these steps may involve deep learning models for object recognition and meshing (Qian et al., 2021; Wang et al., 2021; Li et al., 2021).

### 1.2 SAT2LoD2 software

Due to the implementation triviality and logistics of different libraries, to the authors' knowledge, there are no lightweight open-source tools that often have the capability to reconstruct LoD2 models from Satellite-based products. As a result, researchers either build their flow using a chain of open or commercial tools, with sometimes manual operations, or use open-source tools designed for LiDAR data (Nan & Wonka, 2017; Shah et al., 2017) to reconstruct building models from satellite data. Thus, a compact open-source tool that researchers can perform a "quick and dirty" test, may greatly facilitate researchers who develop relevant methods.

---

*Corresponding author 2036 Neil Avenue, Columbus, Ohio, USA. qin.324@osu.edu

To this end, we developed SAT2LoD2 based on our previously published work (Gui & Qin, 2021), a top-down building model reconstruction approach. The aim is to close this community gap and put these complicated processing steps to drive the development of automated LoD2 modeling approaches. SAT2LoD2 is presented as a GUI-based open-source prototype software based on Python that turns an orthophoto and DSM into individual LoD2 building models. This software has a built-in semantic segmentation module to predict building masks, but can optionally take customized classification maps. It also implements a scheme to utilize road networks (in shapefile) to improve the reconstruction results. During processing, SATR2LoD2 remains all intermediate process output so that users can exploit building models for various purposes. The codes of the proposed method and software are available on GitHub: https://github.com/GDAOSU/LOD2BuildingModel.

## 2. GENERAL ARCHITECTURE AND METHODOLOGY

presents the workflow of our SAT2LoD2 software ( Gui & Qin, 2021). As mentioned earlier, this is a system approach, thus, it consists of a few sequential steps that may all be independently used to serve different geometric processing purposes. The following content introduces the six steps used in our process, where steps 1) and step 6) are updated based on the original work.

1) Building segmentation: generate the building segments from input orthophoto. We adapt HRNet V2 (Sun et al., 2019) as the building segmentation algorithm to extract building segments with high accuracy and efficiency.

2) Initial 2D building polygon extraction: generate the building polygon (boundary) of each building segment, by following initial line extraction, line adjustment, and line refinement.

3) Building rectangle decomposition: generate the building rectangle based on the building polygon using a grid-based decomposition algorithm.

4) Building rectangle orientation refinement: refine the orientation of each building rectangle by combining OpenStreetMap line segments (Haklay, 2010).

5) 3D model fitting: fit and reconstruct the roof structure based on each building rectangle.

6) Mesh export: transform building model into mesh format. For irregular building (low IoU value with mask), directly generate mesh from DSM.

Each step during building model reconstruction can be regarded as an independent function that extracts a specific kind of output result, which can be utilized for purposes other than building generation.

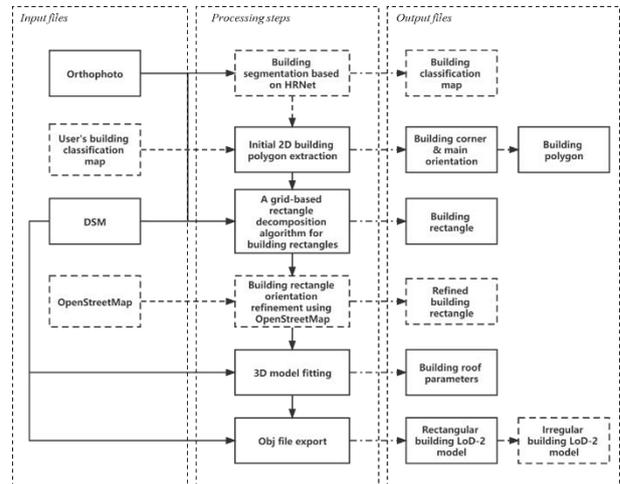

**Fig. 1.** The workflow of our proposed software, and dotted steps are the optional selection.

### 2.1 Building segmentation

The SAT2LoD2 has a trained network that harnesses the existing available benchmark datasets. We used an approach of HRNetV2 (Sun et al., 2019) to get building segments by using orthophoto with RGB bands. The training and validation datasets were combined with satellite and aerial imagery. The satellite dataset was provided by John's Hopkins University Applied Physics Lab (JHUAPL) through the 2019 IEEE GRSS Data Fusion Contest (Le Saux et al., 2019). The aerial dataset was provided by INRIA aerial image labeling benchmark (Maggiori et al., 2017). These datasets have urban buildings from Jacksonville (USA), Omaha (USA), Austin (USA), Chicago (USA), Kitsap (USA), Tyrol (Austria), Vienna (Austria), therefore the trained network is able to work for most of the urban pattern during building segmentation. The images from training and validation datasets are clipped to several $512 \times 512$ patches with 50% overlap. Thus, there will be enough training data with 31615 patches, and 3512 patches for validation.

The training process is a binary classes segmentation in which there are only building and non-building labels for training and validation. During the training step, a total of 30 epochs have been performed with 5000 iterations per epoch, and we adapted the weight from the last epoch for our segmentation function. The training environment is on a GPU of Nvidia RTX 2070s, and the batch size is two during training.

In the prediction part of the software, input orthophoto is divided into $512 \times 512$ patches and then predicted those patches. Moreover, the final segmentation is developed by merging the segment of the predicted individual patch. Thus, the image size processing capability of software depends on the RAM rather than the GPU memory of the user's PC. The output file of this step will be a building segmentation image, which can be used for building detection or building area computation.

### 2.2 Initial 2D building polygon extraction

In order to vectorize building boundaries as polylines, building polygon extraction is applied to regularize building models. For satellite-based data, building masks can be simplified as polylines consisting of parallel or orthogonal lines with three following steps: 1) Initial line extraction gets coarse building boundary by using Douglas–Peucker algorithm (Douglas & Peucker, 1973). 2) Line adjustment process aims to compute main orientations to turn and then connect short line segments to long and straight lines. One or more main orientations can be calculated to separate fitted lines into orientation bins. A given threshold $T_l$ (pixel) decides the length limitation for a 'short' line that can be merged into others. 3) Line regularization process using Line-Segment Detector (LSD) algorithm (Von Gioi et al., 2010) to adjust the orientation of line segments so that they will be consistent with the detected lines from orthophoto. The output file of this step will be the main orientation of each building segment, and building polygon. Researchers can utilize these output files for building footprint extraction without DSM data, or regularize the building shape from satellite images.

### 2.3 A grid-based rectangle decomposition algorithm for building rectangles

It is complicated for polylines from the initial building polygon to fit building models with simplex models. Therefore, these building polygons need to be decomposed into simpler shapes. Our grid-based decomposition approach has an assumption that all complex building polygon consists of several simplex rectangles. The workflow of decomposition can be summarized as 1) rotate the 2D building polygon by aligning its main orientation to the x-axis; 2) initial separate building mask based on the gradient of DSM and orthophoto; 3) build up an image pyramid with three layers and iteratively generate maximum inner rectangles on the coarsest layer, and then interpolated to the finer layers; 4) merge adjacent over-decomposed rectangles with orthophoto and DSM.

In order to decide whether two adjacent rectangles should be merged, the following condition is proposed below:

$$\begin{cases} merge, \ |\overline{C_1} - \overline{C_2}| < T_d \ \cap \ |\overline{H_1} - \overline{H_2}| < T_{h1} \cap \ max|\Delta H_{edge}| < T_{h2} \\ not \ merge, \ otherwise \end{cases} \quad (1)$$

The illustration of threshold $T_d$ (RGB), $T_{h1}$ (m), and $T_{h2}$ (m) are described in **Table. 1**, and these three parameters are the parameters can be tuned in SAT2LoD2 software. The output file of this step is a rectangular building shape, which can be used to mesh generation by including other building modeling methods, or to find the several simple building shapes from a complex and huge building mask.

### 2.4 Orientation refinement for Building rectangles using OpenStreetMap

Noises from the initial building mask and the orthophotos & DSM can easily impact the orientation of decomposed rectangles. At the same time, most neighboring buildings follow consistent orientations, such as those in the same street block. An orientation refinement strategy that detects building rectangles will follow their nearest road vector orientation from the OSM shapefile is adapted to refine the rectangle orientation of buildings near the road, to reduce the bias between decomposed rectangles and road line segments. The output file from this step is refined rectangular building shapes with all 2D parameters, which can be used to analyze the individual building in urban planning research.

### 2.5 3D model fitting

After calculating the building segment's rectangular shape, simplex building models can fit these with DSM. Five types of rectangular building models are considered: flat, gable, hip, pyramid, and mansard.

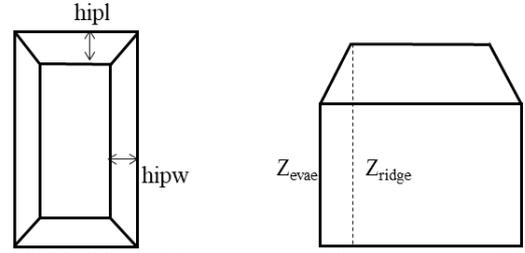

**Fig. 2.** Model roof library with tunable parameters to define these model types

3D geometrical parameters can describe the shape of the building roof, 3D parameters set of $S = \{Z_{ridge}, Z_{eave}, hipl, hipw\}$ can be performed using an exhaustive search to decide the parameter set with the smallest RMSE between DSM, which is illustrated in **Fig. 2**. After this step, 3D parameters and the DSM adapted building model parameters are generated, numerous researches can begin based on those fine building model result.

### 2.6 Model of irregular building shapes

With building models parameters extracted from **Section 2.3 & 2.4** in the 2D level from **Section 2.5** in 3D level, a 3D building mesh can be generated by using these parameters. The processing of converting buildings from DSM to vectorized mesh decreases the redundancy of data, and clean and straightforward geometry with tiny size. However, some buildings may have irregular shapes; in other words, multi-complex models may cause obviously incorrect detections. Therefore, a decision strategy using the Intersection over Union (IoU) based on decomposed rectangles and building masks from segmentation is adapted to discover an irregular shape building model: If IoU between building mask and rectangles in a segment is lower than 0.65, and the building mask has an area larger than 5000 pixels, DSM covered by building polygon in Section 2.2 is transferred into mesh directly, and then mesh is simplified less than 1000 faces. These numbers are tuned specifically for 0.5-meter resolution data. This step ensures that irregular buildings will not be vectorized into an incorrect shape.

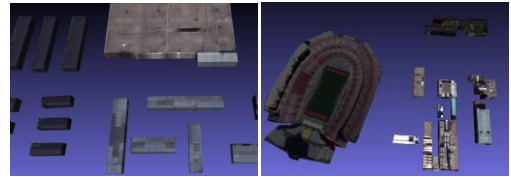

**Fig. 3.** Left: example of the regular (rectangular) building model, right: example of the irregular (non-rectangular) building model

### 3. SOFTWARE MODULES AND OPERATIONS

#### 3.1 Software Architecture

The processing workflow is illustrated in **Section 2**. SAT2LOD2 software is achieved in Python 3.7 environment, and the GUI is designed using PyQt5. The software is generated using the Pyinstaller package to pack up whole Python environments, including Pytorch and other packages. For the architecture of software code, there are eight python files to realize the entire software, other than the GUI file and main function file, and the other six files implement the function in Section 2, respectively. SAT2LoD2 is building LoD-2 reconstruction software that uses typical building 3D model generation pipeline rather than end-to-end deep-learning methods, which deal with the dilemma of no open-source methods available in this field.

There are three parts of input information, 'Input data' indicates the input image file for building model reconstruction; "Reconstruct parameter" represents several significant modeling parameters mentioned in **Section 2.2** and **2.3**; "Output folder" is the path for all output files. If the input data are fulfilled, click on the "OK" button to start processing, and a progress bar will display the current processing status to the user while waiting for the final result.

To ensure that SAT2LoD2 is user-friendly enough for everyone to begin, the only required inputs are orthophoto and DSM for the same area. Building segmentation, building model parameters, and other intermediate results can be generated during the processing. The only additional requirement is that users' PCs need a GPU with a CUDA driver if they plan to use the software's auto segmentation (classification) function. It is still feasible for a PC without GPU to reconstruct the LoD-2 building if the optional input classification map is accessible. Furthermore, a set of default parameters is provided for the user so that the software is easily forthputting enough only to make three selections of orthophoto, DSM, and output folder (if the folder path is empty, the output files will locate at the same path as exclusive file). On the other hand, if the user pursues high accuracy in building a 3D model, they can include their classification, OSM shapefile, and finely tune a set of reconstructing parameters to improve the final performance.

### 3.2 Input data and parameters

For the input images, orthophoto and DSM are required for building model reconstruction, and the classification map and OSM file input can be empty during processing. SAT2LoD2 is flexible, and the format of input files can vary from several different types.

Required input data include: 1) orthophoto with RGB band (.png, .jpg, or .tif format, three bands, 8-bit image); and 2) DSM (digital surface model, .tif format, single band, 16-bit int or 32-bit single image), and the orthophoto and DSM should have the same row and column. The maximum size of orthophoto and DSM is 5000 * 5000 for generating mesh in a few minutes.

Optional input data include: 1) classification image, building class is labeled as 1 or 255 (.png, .jpg, or .tif format with an 8-bit single band, and other types of ground objects need to be set as zeros value); and 2) OpenStreetMap shapefile for road vectors (.shp format), and the corresponding .tfw file, which has the same name as orthophoto, and the projection of both OSM shapefile and .tfw file need to be WGS84 UTM Zone. A typical input file in **Fig. 4** left column contains the orthophoto, DSM, building classification map, and OSM road vector.

There are four important parameters that $T_l$ decides the merged threshold of short lines during polygon extraction, and $T_d$, $T_{h1}$, $T_{h2}$ are the parameter of rectangles merging during grid-based decomposition, which are listed in **Table. 1**. A default set of parameters {$T_l$, $T_d$, $T_{h1}$, $T_{h2}$}={90, 10, 0.5, 0.1} are given to users to easily start. Besides, the range of editing parameters is also given in **Table. 1**, to generate reasonable building models for different areas.

**Table. 1.** Building reconstruction parameters

| Parameter | Description | Range |
|---|---|---|
| Length threshold $T_l$ (pixel) | A threshold of summed up length for determining building main orientations. | [45, 150] |
| Color difference threshold $T_d$ (RGB) | A threshold of mean color differences ($|\overline{C_1} - \overline{C_2}|$) of the two rectangles to decide whether to merge two nearby rectangles in building decomposition. | [6, 20] |
| Mean height difference threshold $T_{h1}$ (m) | A threshold of mean height difference ($|\overline{H_1} - \overline{H_2}|$) between two nearby rectangles to decide whether to merge two nearby rectangles in building decomposition. | [0.5, 1.5] |
| Gap threshold $T_{h2}$ (m) | A threshold of dramatic height changes in a buffered region that cover the common edge between two nearby rectangles. | [0.1, 0.3] |

### 3.3 Output files

All output files will be in the output folder selected in Main Window, and several intermediate process files are generated in this output folder. The building segmentation step will generate the clipped image with a size of 512*512 images with 50% overlap of the orthophoto, and the segmentation result image using HRNet. The initial polygon extraction step will generate the corner, polygon lines, and main orientation for each building segment and visualize the building polygon based on the orthophoto. The building rectangle decomposition step will generate 2D parameters of building rectangles of each building segment and the visualized result. The orientation refinement step will generate building rectangles after refinement and their visualized result. The model-fitting step will generate the 3D parameters of the building roof, and the DSM of that building using the fitted parameters. Finally, the mesh exporting step will generate the LoD-2 mesh file.

Input files of SAT2LoD2 can be only two images, while the output files will be a series of building information. All intermediate results have been generated during processing. Hence, users can also adapt the result other than building mesh to analyze the building information if they are interested. For instance, SAT2LoD can be used to generate a building mask as a building segmentation software, or adapt the building polygon to find the typical shape in one region.

## 4. EXPERIMENTS AND EVALUATION

### 4.1 Example areas

The experiments are performed in three cities with different locations and urban patterns, first one is a typical U.S. city (City of Columbus, Ohio), the other one is located in a European Megacity (London, UK), the last one is in European middle city (Trento, Italy). The orthophoto and DSMs are generated using a multi-view stereo approach (RSP) (Qin, 2016) World-view-2 stereo pairs for the Columbus, London, and Trento dataset. The first two datasets are used to evaluate the performance of building model reconstruction with different kinds of semantic segmentation input. The last dataset is used to evaluate the performance with various quality of DSM input.

The evaluation of the geometry and detection adapts 2D Intersection over Union (IOU2) and 3D Intersection over Union (IOU3) (Kunwar et al., 2020) based on manually created reference data for building footprint and LiDAR-based DSM for 3D geometry.

**4.2 Evaluation of semantic segmentation input**

The input file of semantic segmentation for building class impacts the final accuracy of LoD-2 building model reconstruction. Therefore, the Columbus and London datasets evaluate the quantitative relationship between segmentation results and the building model. The accuracy of the resulting models is evaluated using the IOU2 and IOU3 metrics, respectively, to assess the 2D building footprint extraction accuracy and the 3D fitting accuracy.

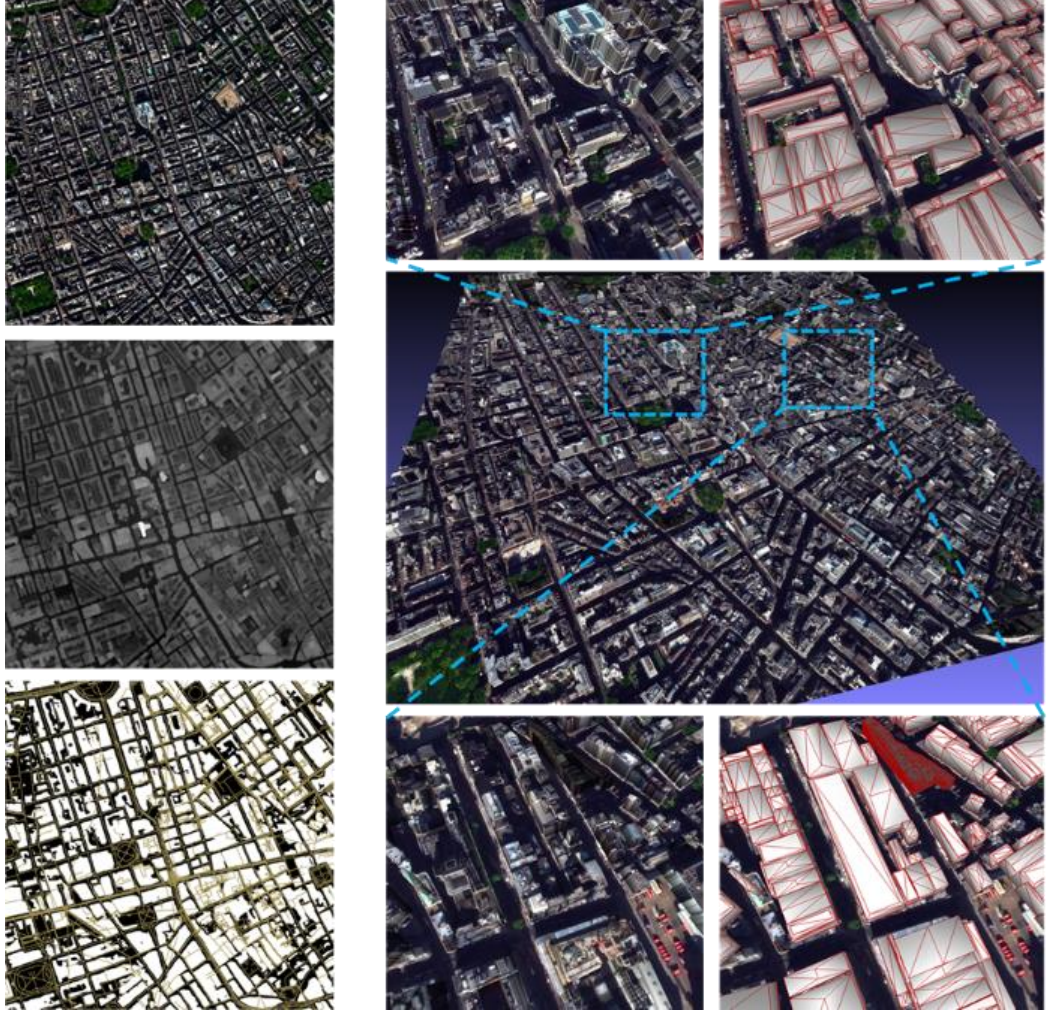

**Fig. 4.** Example result for London-area-1. First column from up to bottom: Orthophoto, DSM, building segmentation map using HRNet, with OSM street lines. Second column: 3D building model and detailed view of two specific areas

HRNet is a capable segmentation model to train and predict building masks, while alternatives such as UNet, Swin-transformer, or other recent semantic segmentation methods can be used as basic building masks to reconstruct building 3D models. We have trained the semantic segmentation network of UNet (Qin et al., 2019), HRNetV2 (Sun et al., 2019), and Swin-transformer (Liu et al., 2021) with the same training dataset from 2019 IEEE GRSS Data Fusion Contest (Le Saux et al., 2019) and INRIA aerial image labeling benchmark (Maggiori et al., 2017). In order to ensure that only input building classification maps have a difference in evaluation, OSM shapefiles are included for all experiments, and the reconstruct parameters adapt the default values of $\{T_l, T_d, T_{h1}, T_{h2}\}$={90, 10, 0.5, 0.1} to generate the building model. **Fig. 5**. indicates the orthophoto and segmentation results from three methods. It can be seen in **Table. 2** that the building model reconstruction accuracy highly depends on the accuracy of the initial building segmentation result.

**Table. 2.** Accuracy evaluation of different segment input

| Region | Accuracy | UNet | HRNet | Swin-T |
|---|---|---|---|---|
| Columbus area 1 | IOU2 (segment) | 0.6927 | 0.7154 | 0.6320 |
| | IOU2 | 0.6310 | 0.5664 | 0.6198 |
| | IOU3 | 0.5636 | 0.5129 | 0.5459 |
| Columbus area 2 | IOU2 (segment) | 0.8355 | 0.8495 | 0.7577 |
| | IOU2 | 0.7984 | 0.7572 | 0.7620 |
| | IOU3 | 0.7836 | 0.7465 | 0.7416 |
| London area 1 | IOU2 (segment) | 0.5866 | 0.6854 | 0.6996 |
| | IOU2 | 0.4019 | 0.5421 | 0.6547 |
| | IOU3 | 0.3578 | 0.4345 | 0.4908 |
| London area 2 | IOU2 (segment) | 0.4856 | 0.6035 | 0.4484 |
| | IOU2 | 0.3845 | 0.4638 | 0.4182 |
| | IOU3 | 0.3128 | 0.4092 | 0.2856 |

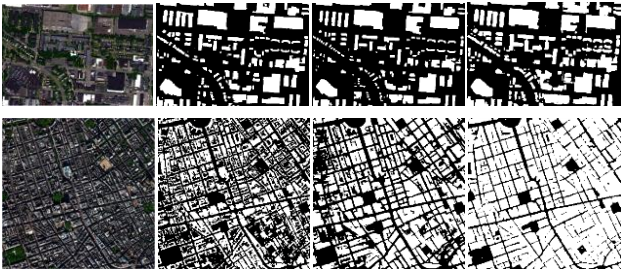

**Fig. 5.** Orthophoto and building segmentation map in example areas of Columbus-area-2 (upper) and London-area-1 (bottom), from left to right: orthophoto, segmentation from UNet, segmentation from HRNet, segmentation from Swin-transformer

### 4.3 Evaluation of DSM input

The input file of DSM impacts the final accuracy of LoD-2 building model reconstruction. Therefore, we adapt the Trento dataset by fusing different amounts of pairs of DSM to generate DSM series with different qualities. The difference of each processing is only input DSM, thus, OSM shapefiles are not used in these cases, and the reconstruct parameters adapt the default values of $\{T_l, T_d, T_{h1}, T_{h2}\}=\{90, 10, 0.5, 0.1\}$ to generate the building model.

DSM series are generated by using one pair to six pairs, and the evaluated result is shown in **Table. 3**. It can be found that with the number of fusion DSM increasing, the fitted model in the IOU2 level follows the improvement as well, and the trend of IOU3 also increases while having a peak when fusing three pairs.

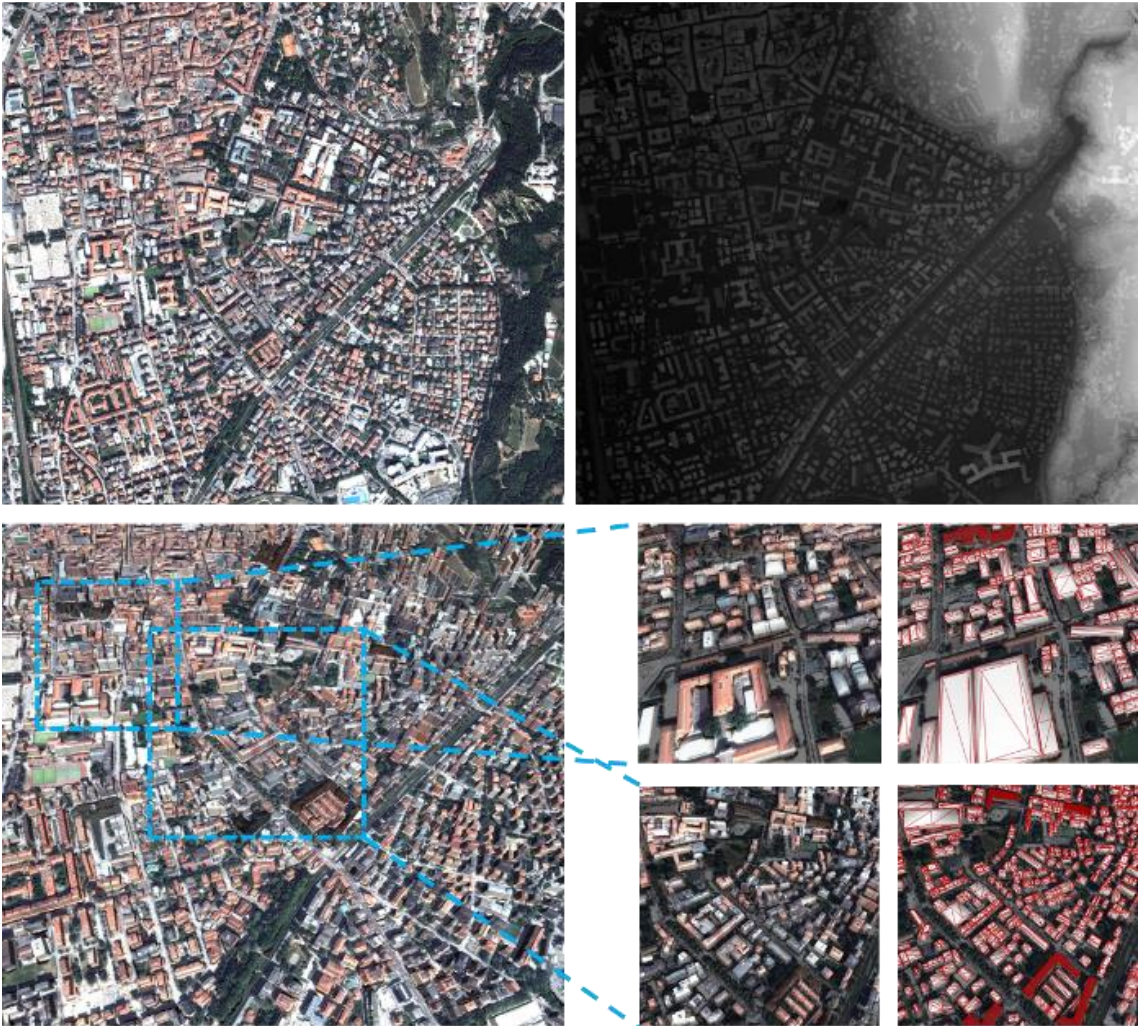

**Fig. 6.** Orthophoto, LiDAR DSM, and generated mesh with one pair of Trento area

**Table. 3.** Accuracy evaluation of different DSM input

| Fused pairs | IOU2 (segment) | IOU2 | IOU3 |
|---|---|---|---|
| 1 pair | | 0.5655 | 0.2505 |
| 2 pairs | 0.6679 | 0.5721 | 0.2772 |
| 3 pairs | | 0.5749 | 0.2995 |
| 4 pairs | | 0.5780 | 0.2901 |

### 4.4 Running time

SAT2LOD is software that combines GPU and CPU during processing, and the GPU part is only used during building segmentation. Therefore, if users have their building mask, the building model can be reconstructed only with the CPU. **Table. 4** shows the specification of testing PC. Since the running time depends on the image size, different datasets are computed, and their running time is shown in **Table. 5**.

**Table. 4.** Specification of the test PC

| CPU | No. of Cores | Memory | GPU |
|---|---|---|---|
| Intel i7-10700k, 5.1 GHz | 8 cores, 16 threads | 32 GB | Nvidia RTX 2070s |

**Table. 5.** Running time on the testing PC (unit: second)

| Areas | Columbus area 1 | Columbus area 2 | London area 1 | London area 2 | Trento area |
|---|---|---|---|---|---|
| Image size (pixel) | 1003*890 | 1646*1118 | 3000*3000 | 3000*3000 | 3680*3309 |
| Segmentation | 12 | 15 | 57 | 55 | 78 |
| Polygon | 6 | 7 | 113 | 127 | 180 |
| Rectangle | 20 | 13 | 110 | 68 | 226 |
| Refinement | 2 | 3 | 30 | 89 | 196 |
| Model fitting | 54 | 30 | 362 | 230 | 215 |
| Obj | 1 | 1 | 22 | 21 | 15 |
| total | 95 | 69 | 694 | 590 | 910 |

## CONCLUSION

This paper introduces an open-source GUI software named SAT2LoD2, which generates building LoD-2 model from satellite-based orthophoto and DSM. This software aims to provide a succinct tool that generates building models robustly, and every researcher can use this tool easily. SAT2LoD2 is able to compute various landscapes in the urban area efficiently. The evaluation of semantic segmentation and DSM input indicates that SAT2LoD2 can provide a convincing result that quality highly depends on the quality of input files. In the next steps, more building models can be included, and processing speed can be further accelerated.

## ACKNOWLEDGEMENT

This work is supported in part by the Office of Naval Research (Award No. N000141712928). Part of the dataset used and involved in this research were relased by IARPA, John Hopkins University Applied Research Lab, IEEE GRSS committee, INRIA, and SpaceNet. The satellite imagery in the MVS benchmark data set was provided courtesy of DigitalGlobe.